\documentclass[lettersize,journal]{IEEEtran}
\usepackage{amsmath,amsfonts}
\usepackage{algorithmic}
\usepackage{algorithm}
\usepackage{array}
\usepackage[caption=false,font=normalsize,labelfont=sf,textfont=sf]{subfig}
\usepackage{textcomp}
\usepackage{stfloats}
\usepackage{url}
\usepackage{verbatim}
\usepackage{graphicx}
\usepackage{cite}
\usepackage{tabularx,siunitx}

\usepackage{multirow}
\usepackage[table,xcdraw]{xcolor}
\usepackage{makecell}

\usepackage{hyperref}

\hypersetup{
    colorlinks = true,
    urlcolor = blue,
    linkcolor = blue,
    citecolor = blue
}

\usepackage{pifont}
\newcommand{\cmark}{\ding{51}}%
\newcommand{\xmark}{\ding{55}}%

\title{AMMNet: An Asymmetric Multi-Modal Network \\ for Remote Sensing Semantic Segmentation}
\author{
Hui~Ye, 
Haodong~Chen, 
Zeke~Zexi~Hu, %
Xiaoming~Chen, 
and Yuk~Ying~Chung,~\IEEEmembership{Member,~IEEE} 

\thanks{
Hui~Ye, Haodong~Chen, Zeke~Zexi~Hu, and Yuk~Ying~Chung are with the School of Computer Science, The University of Sydney, Australia. \\ 
(E-mails: huye0731@uni.sydney.edu.au, haodong.chen@sydney.edu.au, \\ 
zexi.hu@sydney.edu.au, vera.chung@sydney.edu.au,)
}%
\thanks{
Xiaoming~Chen is with the School of Computer and Artificial Intelligence, Beijing Technology and Business University, China. \\ 
(E-mail: xiaoming.chen@btbu.edu.cn)}
}
\date{June 2025}

\begin{document}

\maketitle

\begin{abstract}
Semantic segmentation in remote sensing (RS) has advanced significantly with the incorporation of multi-modal data, particularly the integration of RGB imagery and the Digital Surface Model (DSM), which provides complementary contextual and structural information about the ground object.
However, integrating RGB and DSM often faces two major limitations: increased computational complexity due to architectural redundancy, and degraded segmentation performance caused by modality misalignment.
These issues undermine the efficiency and robustness of semantic segmentation, particularly in complex urban environments where precise multi-modal integration is essential.
To overcome these limitations, we propose Asymmetric Multi-Modal Network (AMMNet), a novel asymmetric architecture that achieves robust and efficient semantic segmentation through three designs tailored for RGB-DSM input pairs.
To reduce architectural redundancy, the Asymmetric Dual Encoder (ADE) module assigns representational capacity based on modality-specific characteristics, employing a deeper encoder for RGB imagery to capture rich contextual information and a lightweight encoder for DSM to extract sparse structural features.
Besides, to facilitate modality alignment, the Asymmetric Prior Fuser (APF) integrates a modality-aware prior matrix into the fusion process, enabling the generation of structure-aware contextual features.
Additionally, the Distribution Alignment (DA) module enhances cross-modal compatibility by aligning feature distributions through divergence minimization.
Extensive experiments on the ISPRS Vaihingen and Potsdam datasets demonstrate that AMMNet attains state-of-the-art segmentation accuracy among multi-modal networks while reducing computational and memory requirements. 
These results validate the effectiveness and robustness of the proposed asymmetric design for multi-modal semantic segmentation in complex urban environments.

\end{abstract}


\section{Introduction}
\IEEEPARstart{R}{emote} sensing semantic segmentation is a fundamental task in geoscientific research, enabling the transformation of raw multispectral, hyperspectral, or LiDAR data into structured spatial representations.
By generating pixel-level classification of land cover and land use, this process supports accurate geospatial analysis and serves as a foundational tool for numerous remote sensing applications, such as environmental monitoring~\cite{Asadzadeh2022, Yuan2020}, resource management~\cite{Zhai2020}, and disaster assessment~\cite{Kakooei2022}.

The increasing complexity of Earth observation missions and the growing demand for fine-grained analysis have driven the advancement of remote sensing semantic segmentation techniques.
Especially, recent progress in deep neural networks has enabled high-resolution, task-specific perception across a wide range of remote sensing applications.
However, the robustness of deep neural networks relying on single-modal input is often compromised under complex remote sensing (RS) conditions, including illumination changes, sensor noise, occlusions, and seasonal variations.

\begin{figure}[t]
    \centering
    \includegraphics[width=.95\linewidth]{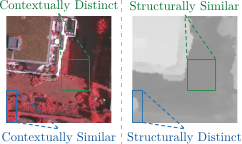}
    \caption{Visualization of the complementary characteristics between RGB and DSM modalities. Both RGB and DSM provide distinct yet complementary information, where each modality compensates for the ambiguities or limitations of the other.}
    \label{fig:input_vis}
\end{figure}

Multi-modal learning, which refers to the integration of heterogeneous data sources for a specific task, is inherently well-suited for RS due to the availability of diverse data modalities such as RGB imagery and Digital Surface Models (DSM).
An example of these input modalities is illustrated in Fig.~\ref{fig:input_vis}.
RGB imagery, typically derived from the visible portion of the electromagnetic spectrum, primarily captures contextual information~\cite{Chen2024c}. 
However, it is highly susceptible to external disturbances such as cloud cover, fog, and atmospheric aerosols~\cite{Xu2023, Du2024a}, which can induce intra-class variability and inter-class similarity~\cite{Luo2024}, thereby posing challenges to robust semantic segmentation.
In contrast, Digital Surface Models (DSM) provide elevation-based structural information that, although lacking rich semantic content, offers a distinct and complementary perspective to RGB imagery. 
This structural cue is particularly valuable for distinguishing objects with high contextual similarity but varying physical heights, such as roads versus buildings or low vegetation versus trees.

Although previous studies~\cite{Ma2022, Ma2024} have demonstrated the promising potential of multi-modality semantic segmentation, several critical limitations still hinder its effectiveness.
One of the limitations is architectural redundancy~\cite{Bao2025}, wherein the inclusion of extra encoder and fusion components leads to increased computational complexity. 
Another fundamental issue is modality misalignment~\cite{Bao2025, Luo2024}, which stems from the inherent heterogeneity between modalities.
If not properly addressed, such misalignment can introduce irrelevant or conflicting information, leading to performance degradation.

To address these limitations, we propose the Asymmetric Multi-Modal Network (AMMNet), which incorporates three asymmetric modules.
To reduce architectural redundancy, we adopt an asymmetric encoder design, termed the Asymmetric Dual Encoder (ADE), which employs modality-specific encoders tailored to the distinct characteristics of each modality. 
Specifically, the RGB branch is assigned a greater capacity to extract dense contextual information, while the DSM branch uses a lightweight encoder to extract sparse structural information.
In addition, two asymmetric designs are introduced to address modality misalignment:
The Asymmetric Prior Fuser (APF) integrates RGB and DSM features through a modality-aware prior matrix, generating structure-aware contextual representation.
The Distribution Alignment (DA) module further minimizes the distributional divergence between the DSM and RGB features, enhancing cross-modal compatibility.

Therefore, the main contributions of this work are summarized as follows:
\begin{itemize}
    \item To achieve efficient and robust semantic segmentation in a complex urban environment, we propose AMMNet, a novel multi-modal network that employs modality-specific asymmetric designs to effectively leverage RGB-DSM modality pairs.

    \item We introduce the Asymmetric Dual Encoder (ADE), which employs modality-specific encoder configurations tailored to the distinct characteristics of RGB-DSM inputs, enabling reduced computational complexity while maintaining representational capacity.
    
    \item To address modality misalignment, we propose two asymmetric modules. The Asymmetric Prior Fuser (APF) constructs a modality-aware prior matrix to generate structure-aware contextual features, while the Distribution Alignment (DA) module enhances cross-modal compatibility between RGB and DSM inputs.

    \item Extensive experiments on two benchmark datasets, ISPRS Vaihingen and Potsdam, demonstrate that AMMNet achieves superior semantic segmentation performance in complex urban environments while maintaining computational efficiency, outperforming advanced multi-modal approaches.
\end{itemize}

\section{Related Work}
\subsection{Remote Sensing Semantic Segmentation}
With the rapid advancement of generic semantic segmentation techniques, network architecture has evolved from convolutional neural networks (CNNs)~\cite{Long2015, Ronneberger2015, Zhao2017, Chen2018a, Chen2018b} to transformer-based designs~\cite{Xie2021, Liu2021a}.
More recently, large vision models (VLM)~\cite{Kirillov2023} have emerged as a promising paradigm, potentially defining the next generation of segmentation architectures.

However, remote sensing semantic segmentation continues to encounter some domain-specific challenges, such as ultra-high spatial resolution~\cite{Huang2025, Luo2024}, seasonal variations~\cite{Zhang2023, Yang2022}, cloud occlusion~\cite{Xu2023, Du2024a}, and so on.
Numerous advanced techniques have been proposed to enhance the semantic segmentation performance, including global-local feature integration~\cite{Cao2023, Wang2022}, multi-scale pyramid structures~\cite{Li2022}, and the adaptation of VLMs~\cite{Wang2023a, Ma2024a, Chen2024a}. 
It is worth noting that most remote sensing semantic segmentation methods still rely on single-modal input, primarily RGB, which inherently leads to limited robustness due to the restricted information available from a single sensor.

\subsection{Multi-modal Learning in Remote Sensing}
Multi-modal learning has emerged as a promising paradigm, as it enables the integration of complementary information from multiple sensors, thereby enhancing performance on downstream tasks.
It is widely adopted across a range of domains, such as natural image processing~\cite{Zhu2025}, medical imaging~\cite{Chen2024}, and autonomous driving~\cite{Seichter2022, zhang2023b, zhang2023a, Caltagirone2019}.
The widespread deployment underscores its versatility and growing significance in advancing visual perception across diverse and complex real-world scenarios.

Multi-modal learning is intuitively well-suited for remote sensing due to the availability of diverse modalities, such as hyperspectral imagery (HSI), digital surface models (DSM), synthetic aperture radar (SAR), and panchromatic (PAN) imagery.
Therefore, it has been widely employed in a variety of remote sensing tasks, including pansharpening~\cite{Huang2025}, object detection~\cite{Zhao2025, Bao2025, Wang2024a, Zhang2024a}, change detection~\cite{Zhang2023, Yang2022, chen2024b}, cloud removal~\cite{Xu2023, Du2024a}, and semantic segmentation~\cite{Zhao2021, Cui2024, Ma2022, Ma2024, Sun2022, Wang2024}.

\subsection{Asymmetric Architecture}
Asymmetric architectures are specifically designed for multi-modal learning to address the inherent heterogeneity among input modalities. 
Modalities, such as RGB, DSM, SAR, and HSI, differ significantly in spectral content, spatial resolution, and structural characteristics. 
In contrast, symmetric architectures, which apply uniform processing across modalities, often result in redundant computation or insufficient representational learning, failing to capture the unique properties of each modality effectively.

As deep neural networks continue to scale in size and complexity, the risk of architectural redundancy becomes more pronounced, increasing both training and inference costs. 
The asymmetric design can be a promising solution to this challenge by allocating computational resources based on the information density of each modality, thereby reducing architectural redundancy and computational overhead~\cite{Liu2024}.

Recent works have demonstrated the effectiveness of asymmetric architectures across various granularities, including layer-level~\cite{Fan2024}, attention-level~\cite{Zhao2023a}, and branch-level~\cite{Liu2024} configurations. 
In particular, this strategy has been successfully applied to a range of RS applications, such as haze removal~\cite{Xu2023, Du2024a}, change detection~\cite{Zhang2023, Yang2022}, pansharpening~\cite{Zhao2023a}, and semantic segmentation~\cite{Liu2024, Fan2024, Chan2023}. 
These studies highlight that well-designed asymmetric architecture enables more effective integration of complementary modality information, thereby improving performance while maintaining efficiency.

\section{Methodology}
This section details the proposed AMMNet framework.
Section~\ref{sec:overview} provides an overall architectural overview.
Section~\ref{sec:encoder} introduces the Asymmetric Dual Encoder (ADE), which reduces architectural redundancy while preserving representational capacity.
Sections~\ref{sec:fuser} and~\ref{sec:alignment} discuss how modality misalignment can be mitigated via modality-aware fusion and distribution-level alignment, respectively. 
Finally, Section~\ref{sec:loss_fn} defines the training objectives and associated loss functions.

\subsection{Overview}
\label{sec:overview}

\begin{figure*}[!htp]
    \centering
    \includegraphics[width=\linewidth]{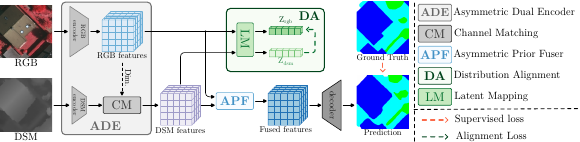}
    \caption{The overall architecture of AMMNet. The input RGB-DSM pair is first processed by the Asymmetric Dual Encoder (ADE), which employs a deep-shallow encoder configuration, along with a Channel Matching (CM) module. Subsequently, the Asymmetric Prior Fuser (APF) integrated RGB and DSM features to generate fused features. In parallel, the Distribution Alignment (DA) module enhances cross-modal compatibility between RGB and DSM features. Finally, the fused features are decoded via a decoder to produce the final semantic segmentation map.}
    \label{fig:ammnet}
\end{figure*}

An overview of the proposed AMMNet network is illustrated in Fig.~\ref{fig:ammnet}.
The input RGB-DSM pairs are first processed by the Asymmetric Dual Encoder (ADE), which assigns a deep encoder to the RGB modality and a shallow encoder to the DSM modality for efficient and modality-specific feature extraction.
To address the resulting dimension mismatch, a Channel Matching (CM) module is incorporated into ADE to match DSM features with their RGB counterparts.
Despite introducing minimal additional parameters, CM allows ADE to maintain strong representational capability while reducing overall architectural redundancy.

Then, the RGB and DSM features extracted by ADE are integrated by the Asymmetric Prior Fuser (APF). 
Unlike symmetric fusion strategies, APF constructs a modality-aware prior matrix that captures complementary information from both modalities, facilitating more effective feature integration.
This prior is subsequently incorporated into the RGB contextual features to generate structure-aware contextual features.
Concurrently, the features from ADE are aligned by the Distribution Alignment (DA) module, which projects them into a latent space via Latent Mapping (LM) to capture their underlying distributions.
By minimizing the divergence between the latent variables of RGB and DSM, the DA module ensures cross-modal compatibility.

Finally, the fused features are fed into a decoder adapted from~\cite{Wang2022}, which integrates both global and local contextual information to generate the final semantic segmentation maps.

\subsection{Asymmetric Dual Encoder}
\label{sec:encoder}

\begin{figure}[!tp]
    \centering
    \includegraphics[width=\linewidth]{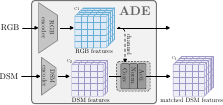}
    \caption{The overview of Asymmetric Dual encoder (ADE). It first extracts features from RGB and DSM input using modality-specific encoders. The Channel Matching (CM) module then aligns the DSM features dimension with those of the RGB features to ensure consistency for downstream integration.}
    \label{fig:ADE}
\end{figure}

Symmetric architectures typically adopt identical encoder designs for both RGB and DSM modalities. However, such a design may result in redundant computation or suboptimal representation, as it ignores the inherent differences between modalities. 
RGB images provide rich contextual information, necessitating a deep encoder with strong representational capacity. 
In contrast, DSM data primarily encodes sparse structural cues, for which a shallow encoder is sufficient. 
Accordingly, AMMNet adopts an asymmetric encoder configuration, termed the Asymmetric Dual Encoder (ADE), as illustrated in Fig.~\ref{fig:ADE}, which improves modality-specific feature extraction while reducing computational overhead.

Nevertheless, this asymmetric encoder configuration introduces a channel dimensionality mismatch: RGB features typically possess a higher channel dimension $c_1$ to capture rich contextual information, while DSM features have a lower channel dimension $c_2$, reflecting their sparse structural content.
To address this, the Channel Matching (CM) module projects DSM features from $c_2$ to $c_1$, matching them with the RGB features.
This direction of matching is chosen to preserve the rich information encoded in the RGB features without compressing or distorting them.
The CM comprises a $1 \times 1$ convolution followed by batch normalization and ReLU activation, enabling efficient dimensional adjustment with minimal overhead.
With CM, ADE generates dimensionally compatible feature pairs, maintaining representational strength with less computational cost.

\subsection{Asymmetric Prior Fuser}
\label{sec:fuser}

\begin{figure}[!tp]
    \centering
    \includegraphics[width=\linewidth]{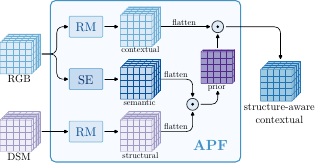}
    \caption{The overview of Asymmetric Prior Fuser (APF). Two Residual Mappers (RM) are employed to extract contextual and structural features from RGB and DSM modalities, respectively. A Semantic Enhancer (SE) is applied to the RGB branch to retain its semantic features. A prior matrix is then constructed by combining the semantic and structural features. This prior is subsequently fused with the contextual features to produce a final representation that embeds structure-aware contextual information.}
    \label{fig:apf}
\end{figure}

Due to the difference in sensor measurement techniques, RGB and DSM modalities often exhibit misalignment, posing challenges for effective modality integration.
Such misalignment can introduce irrelevant or even contradictory information, potentially degrading the representational quality of the fused features.

Unlike symmetric fusion approaches~\cite{Hosseinpour2022, He2023a, Ma2022, Chen2020, Chen2024, Seichter2022, Hazirbas2017, Audebert2018, Ma2024}, which typically extract partial information from one modality and inject it into another, our Asymmetric Prior Fuser (APF), as illustrated in Fig.~\ref{fig:apf}, leverages an asymmetric design to generate structure-aware contextual features by integrate RGB-DSM input pairs with a modality-aware prior.

To preserve modality-specific information, we employ two Residual Mapping (RM) modules: one for the RGB modality and one for the DSM modality.
Each RM module retains contextual features ${f_{con}}$ and structural features ${f_{str}}$ from their respective modalities via a linear layer.
In addition, a Semantic Enhancer (SE) module, comprising a linear layer followed by batch normalization and ReLU activation, is asymmetrically applied only to the RGB branch to extract semantic features ${f_{sem}}$.

Compared to the general contextual features ${f_{con}}$, semantic features ${f_{sem}}$ tend to exhibit stronger alignment with structural features ${f_{str}}$, as both capture object-level cues such as boundaries, shapes, and spatial layouts. 
To leverage this alignment, we compute a prior matrix as follows:
\begin{equation}
    f_{prior} = \text{softmax} (\frac{f_{sem} \odot f_{str}^\top}{\sqrt{d_{str}}}),
\end{equation}
\noindent
where $d_{\text{str}}$ denotes the feature dimension of $f_{\text{str}}$.
This prior serves as a modality-aware representation that is consistent with both RGB and DSM modalities.

Finally, the prior matrix, constructed from the interaction between semantic features and structural features, is used to refine the contextual features, yielding the fused feature $f_{fuse}$ that embeds structure-aware contextual information:
\begin{equation}
    f_{fuse} = f_{prior} \odot f_{con},
\end{equation}
\noindent
which reflects an asymmetric fusion strategy, where guidance is derived from one set of cross-modal features and applied selectively to another, allowing the model to exploit complementary modality characteristics without homogenizing their roles.

In summary, the APF module employs an asymmetric fusion design to generate structure-aware contextual features. 
This is achieved by constructing a modality-aware prior that is tailored to the inherent characteristics of the RGB and DSM modalities, effectively leveraging their complementary strengths.

\subsection{Distribution Alignment}
\label{sec:alignment}

\begin{figure}[!tp]
    \centering
    \includegraphics[width=\linewidth]{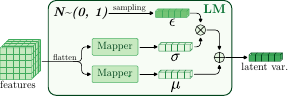}
    \caption{The overview of the Latent Mapping (LM) module. The input features are first flattened and passed through two mappers to generate the mean $\mu$ and standard deviation $\sigma$. A latent variable is then sampled by applying the reparameterization trick, where a random variable $\epsilon$ is drawn from a standard Gaussian distribution $\mathcal{N}(0, 1)$ and scaled by $\sigma$, then shifted by $\mu$: $z = \mu + \sigma \cdot \epsilon$.}
    \label{fig:reparam}
\end{figure}

Due to their distinct measurement techniques, RGB and DSM modalities exhibit different data distributions, which may lead to semantic ambiguity and potentially introduce irrelevant or even contradictory information.

To mitigate the ambiguity arising from distributional discrepancies, we propose the Distribution Alignment (DA) module, which minimizes the divergence between the feature distribution of DSM and RGB. 

To perform this divergence minimization, DA firstly maps features into latent variables, which represent their underlying distributions, via the Latent Mapping (LM) module, as illustrated in Fig.~\ref{fig:reparam}.
The input features are first flattened along the channel dimension and passed through two linear mappers to produce two vectors of length $L$, representing the mean $\mu$ and standard deviation $\sigma$, respectively.
The latent variables $z$ are then computed using the reparameterization trick~\cite{Kingma2013}, where a noise vector $\epsilon$ is sampled from a standard Gaussian distribution:
\begin{equation}
     z = \mu + \sigma \cdot \epsilon.
\end{equation}

Let $z_{\text{rgb}}$ and $z_{\text{dsm}}$ denote the latent variables of RGB and DSM, respectively. we align $z_{dsm}$ to $z_{rgb}$ by minimizing the following alignment loss:
\begin{equation}
    \mathcal{L}_{\text{align}} = \sum_{i=1}^{D} p_i^{\text{dsm}} \left( \log p_i^{\text{dsm}} - \log p_i^{\text{rgb}} \right),
\label{eq:align_loss}
\end{equation}
\noindent
where $p_{i}^{\text{rgb}}$ and $p_{i}^{\text{dsm}}$ denote the probability derived from the RGB and DSM latent variables, $z_i^{rgb}$ and $z_i^{dsm}$, respectively. 

This asymmetric design facilitates cross-modal compatibility by preserving complementary modality-specific information while suppressing redundant and task-irrelevant content.
This capability is particularly important given that the RGB modality provides richer contextual information, which plays a dominant role in semantic segmentation.

\subsection{Loss Function}
\label{sec:loss_fn}
The primary objective of the segmentation task is to minimize the supervised loss $\mathcal{L}_{sup}$, defined as:
\begin{equation}
\mathcal{L}_{\text{sup}} = -\frac{1}{N} \sum_{i=1}^N \sum_{c=1}^C y_{i,c} \log(\hat{y}_{i,c}),
\label{eq:sup}
\end{equation}
\noindent 
where $N$ is the number of samples, $C$ is the number of classes, $y_{i,c}$ denotes the one-hot encoded ground-truth label for class $c$ at sample $i$, and $\hat{y}_{i,c}$ the predicted probability for class $c$ at sample $i$.

To achieve cross-modal compatibility, an alignment loss, denoted as $\mathcal{L}_{\text{align}}$ and defined in Eq.~\ref{eq:align_loss}, is incorporated into the overall training objective.
Accordingly, the final objective function combines supervised learning with distribution alignment, is formulated as:
\begin{equation}
    \mathcal{L_{\text{final}}} = \mathcal{L}_{sup} + \alpha \cdot \mathcal{L}_{align},
\label{eq:final_loss}
\end{equation}
\noindent
where $\alpha$ is a weighting coefficient that controls the contribution of the alignment loss.

\section{Experiment}
\subsection{Datasets}
\subsubsection{ISPRS Vaihingen}
The ISPRS Vaihingen dataset, provided by the International Society for Photogrammetry and Remote Sensing (ISPRS), is a widely used benchmark for semantic segmentation in remote sensing. 
It comprises 33 high-resolution true orthophoto (TOP) images captured over Vaihingen, Germany.
Each image includes three spectral bands (Infrared, Red, and Green), along with an associated digital surface model (DSM). 
The spatial resolution of the images ranges from $1996 \times 1995$ to $3816 \times 2550$ pixels.
Pixel-level annotations are provided for six semantic classes: Building (Bui.), Tree (Tre.), Low vegetation (Low.), Car, Impervious surface (Imp.), and clutter (background).

Following the experimental setup in~\cite{Ma2024}, 12 images are used for training and 4 images for testing. 
For single-modal networks, only IRRG images are utilized, while multi-modal networks utilize both IRRG and DSM data. 
The training set comprises images with the following IDs: 1, 3, 23, 26, 7, 11, 13, 28, 17, 32, 34, and 37. 
The test set includes images with IDs: 5, 21, 15, and 30.

\subsubsection{ISPRS Potsdam}
The ISPRS Potsdam dataset~\cite{ISPRS_potsdam} is another high-resolution benchmark provided by ISPRS for urban semantic segmentation. 
It includes 38 TOP images captured over Potsdam, Germany, each containing 4 spectral channels (Red, Green, Blue, and Near-Infrared), along with corresponding DSM data. 
Unlike the Vaihingen dataset, all images have a uniform spatial resolution of $6000 \times 6000$ pixels. 
Semantic annotations are provided for the same six classes as in the Vaihingen dataset: Building (Bui.), Tree (Tre.), Low vegetation (Low.), Car, Impervious surface (Imp.), and Clutter (Background).

Consistent with~\cite{Ma2024}, we use 18 images for training and 6 for testing. 
Single-modality networks utilize only RGB images, whereas multi-modal networks leverage both RGB and DSM data. 
The training set includes images with IDs: $2\_10$, $2\_12$, $3\_11$, $3\_12$, $4\_11$, $4\_12$, $5\_10$, $5\_12$, $6\_7$, $6\_8$, $6\_9$, $6\_10$, $6\_12$, $7\_7$, $7\_8$, $7\_9$, $7\_10$, and $7\_11$. 
The test set comprises IDs: $2\_11$, $3\_10$, $4\_10$, $5\_11$, $6\_11$, and $7\_12$.

\subsection{Implementation Detail}
All networks are implemented using the PyTorch framework and trained on an NVIDIA GTX 4080 GPU with 16 GB of memory. 
To optimize training efficiency, the AdamW optimizer is employed in all experiments. 
The base learning rate is set to $2 \times 10^{-4}$, with a cosine annealing schedule applied for dynamic adjustment during training.
Swin Transformer encoder \cite{Liu2021a} is employed for the asymmetric dual encoder.

For the Vaihingen and Potsdam datasets, input images are randomly cropped into $256 \times 256$ patches. 
To enhance network generalization, data augmentation techniques, including random resizing, vertical and horizontal flips, and random rotations, are applied during training. 
Each network is trained for 100 epochs with a batch size of 8.

\subsection{Evaluation Metrics}
The evaluation of our experiments is conducted using two primary categories of metrics.

The first category focused on measuring the accuracy of networks, including mean overall accuracy (mOA), mean F1 score (mF1), and mean intersection over union (mIoU).

The mean overall accuracy (mOA) quantifies the proportion of correctly classified pixels across the dataset and is defined as:
\begin{equation}
    \text{mOA} = \frac{1}{K} \sum_{k=1}^{K} \frac{n_{kk}}{t_k},
\end{equation}
\noindent
where $K$ is the total number of classes, $n_{kk}$ is the number of true positives for class $k$, and $t_k$ is the total number of ground truth pixels of class $k$.

The mean F1 score (mF1) represents the harmonic mean of precision and recall, averaged across all classes. 
It is given by:
\begin{equation}
    \text{mF1} = \frac{1}{K} \sum_{k=1}^{K} \frac{2n_{kk}}{p_k + t_k},
\end{equation}
\noindent
where $p_k$ is the total number of pixels predicted as class $k$, and $t_k$ is the total number of ground truth pixels of class $k$.

The mean Intersection over Union (mIoU) measures the average overlap between predicted and ground truth regions for all classes and is formulated as:
\begin{equation}
    \text{mIoU} = \frac{1}{K} \sum_{k=1}^{K} \frac{n_{kk}}{t_k + p_k - n_{kk}},
\end{equation}
\noindent
where $n_{kk}$ denotes the intersection between the predicted and ground truth pixels of class $k$, and $t_k + p_k - n_{kk}$ represents the corresponding union.

The second category evaluates the efficiency of the networks. 
This includes the number of floating-point operations (FLOPs), which quantifies computational complexity by counting the total number of operations required during inference; the memory footprint (MB), which measures the memory consumption; and the number of model parameters (M), which reflects the overall model size.

\subsection{Quantitative Comparison}
The proposed AMMNet is evaluated against 14 advanced networks, including 5 single-modal networks and 9 multi-modal networks, on two benchmark datasets, ISPRS Vaihingen and Potsdam.

\subsubsection{Performance Comparison on Vaihingen Dataset}

\begin{table*}[!t]
\tiny
\centering
\caption{Comparative result on Vaihingen dataset. Best values are highlighted in Bold. Category-wise results are presented using mean F1 scores.}
\label{tab:result_vaihingen}
\resizebox{\linewidth}{!}{%
\begin{tabular}{c|c|ccccc|ccc}
\hline
\textbf{Modality} & \textbf{Model} & \textbf{Bui.} & \textbf{Tre.} & \textbf{Low.} & \textbf{Car} & \textbf{Imp.} & \textbf{mOA} $\uparrow$ & \textbf{mF1} $\uparrow$ & \textbf{mIoU} $\uparrow$ \\ \hline
 & ABCNet~\cite{Liu2021} & 94.10 & 90.81 & 78.53 & 64.12 & 89.70 & 89.25 & 85.34 & 75.20 \\
 & PSPNet~\cite{Zhao2017} & 94.52 & 90.17 & 78.84 & 79.22 & 92.03 & 89.94 & 86.55 & 76.96 \\
 & MAResU-Net~\cite{Li2022} & 94.84 & 89.99 & 79.09 & 85.89 & 92.19 & 90.17 & 88.54 & 79.89 \\
 & Swin-UNet~\cite{Cao2023} & 95.39 & 91.79 & 77.18 & 70.01 & 90.42 & 89.91 & 85.85 & 76.03 \\
\multirow{-5}{*}{Single} & UNetFormer~\cite{Wang2022} & 96.23 & 91.85 & 79.95 & 86.99 & 91.85 & 91.17 & 89.85 & 81.97 \\ \hline
 & vFuseNet~\cite{Audebert2018} & 95.92 & 91.36 & 77.64 & 76.06 & 91.85 & 90.49 & 87.89 & 78.92 \\
 & FuseNet~\cite{Hazirbas2017} & 96.28 & 90.28 & 78.98 & 81.37 & 91.66 & 90.51 & 87.71 & 78.71 \\
 & ESANet~\cite{Seichter2022} & 95.69 & 90.50 & 77.16 & {\color[HTML]{000000} 85.46} & 91.39 & 90.61 & 88.18 & 79.42 \\
 & SA-GATE~\cite{Chen2020} & 94.84 & 92.56 & 81.29 & 87.79 & 91.69 & 91.10 & 89.81 & 81.27 \\
 & CMGFNet~\cite{Hosseinpour2022} & 97.75 & 91.60 & 80.03 & 87.28 & 92.35 & 91.72 & 90.00 & 82.26 \\
 & TransUNet~\cite{Chen2024} & 96.48 & 92.77 & 76.14 & 69.56 & 91.66 & 90.96 & 87.34 & 78.26 \\
 & CMFNet~\cite{Ma2022} & 97.17 & 90.82 & 80.37 & 85.47 & 92.36 & 91.40 & 89.48 & 81.44 \\
 & MFTransNet~\cite{He2023a} & 96.41 & 91.48 & 80.09 & 86.52 & 92.11 & 91.22 & 89.62 & 81.61 \\
 & FTransUNet~\cite{Ma2024} & \textbf{98.20} & 91.94 & 81.49 & 91.27 & 93.01 & 92.40 & 91.21 & 84.23 \\
\multirow{-10}{*}{Multiple} & \cellcolor[HTML]{D3D3D3} AMMNet (Ours) & \cellcolor[HTML]{D3D3D3}{96.06} & \cellcolor[HTML]{D3D3D3}{\textbf{96.55}} & \cellcolor[HTML]{D3D3D3}{\textbf{86.83}} & \cellcolor[HTML]{D3D3D3}{\textbf{93.91}} & \cellcolor[HTML]{D3D3D3}{\textbf{94.24}} & \cellcolor[HTML]{D3D3D3}{\textbf{93.52}} & \cellcolor[HTML]{D3D3D3}{\textbf{93.27}} & \cellcolor[HTML]{D3D3D3}{\textbf{87.56}} \\ \hline
\end{tabular}%
}
\end{table*}

As shown in Table~\ref{tab:result_vaihingen}, AMMNet achieves state-of-the-art (SOTA) performance on the Vaihingen dataset across all evaluation metrics, including an mOA of 93.52\%, mF1 of 93.27\%, and mIoU of 87.56\%.

The results highlight the general advantages of multi-modal networks over single-modal counterparts, primarily due to their ability to incorporate structural information from the DSM modality.
However, not all multi-modal networks (vFuseNet, FuseNet, and ESAnet) outperform single-modal networks such as UNetformer, likely due to the inappropriate fusion strategies that introduce irrelevant or contradictory information.

AMMNet also surpasses other multi-modal networks, owing to its innovative asymmetric design that enables more effective and targeted integration of contextual and structural information, avoiding the limitation of symmetric strategies. 
Notably, AMMNet achieves substantial improvements of 4.61\% and 5.34\% in the ``Tree'' and ``Low Vegetation'' categories, respectively, compared to the previous SOTA network, FTransUNet, highlighting its effectiveness in handling structurally complex and contextually ambiguous classes. 
In addition, it achieves overall performance gains of 3.33\% in mIoU, 2.06\% in mF1, and 1.12\% in mOA.

\subsubsection{Performance Comparison on Potsdam Dataset}

\begin{table*}[!t]
\tiny
\centering
\caption{Comparative result on Potsdam dataset. Best values are highlighted in Bold. Category-wise results are presented using mean F1 scores.}
\label{tab:result_potsdam}
\resizebox{\textwidth}{!}{%
\begin{tabular}{c|c|ccccc|ccc}
\hline
\textbf{Modality} & \textbf{Models} & \textbf{Bui.} & \textbf{Tre.} & \textbf{Low.} & \textbf{Car} & \textbf{Imp.} & \textbf{mOA} $\uparrow$ & \textbf{mF1} $\uparrow$ & \textbf{mIoU} $\uparrow$ \\ \hline
& ABCNet~\cite{Liu2021} & 96.23 & 78.92 & 86.40 & 92.92 & 88.90 & 87.52 & 88.14 & 79.26 \\
& PSPNet~\cite{Zhao2017} & 97.03 & 83.13 & 85.67 & 88.81 & 90.91 & 88.67 & 88.92 & 80.36 \\
& MAResU-Net~\cite{Li2022} & 96.82 & 83.97 & 87.70 & 95.88 & 92.19 & 89.82 & 90.86 & 83.61 \\
& Swin-UNet~\cite{Cao2023} & 97.45 & 86.77 & 87.92 & 94.23 & 90.10 & 90.17 & 91.33 & 84.28 \\
\multirow{-5}{*}{Single} & UNetFormer~\cite{Wang2022} & 97.69 & 86.47 & 87.93 & 95.91 & 92.27 & 90.65 & 90.71 & 85.05 \\ \hline
& vFuseNet~\cite{Audebert2018} & 97.48 & 85.14 & 87.31 & 96.10 & 92.64 & 90.58 & 91.60 & 84.86 \\
& FuseNet~\cite{Hazirbas2017} & 97.10 & 85.31 & 87.81 & 94.08 & 92.76 & 89.74 & 91.22 & 84.15 \\
& ESANet~\cite{Seichter2022} & 96.54 & 81.18 & 85.35 & \textbf{96.63} & 90.77 & 87.91 & 90.26 & 82.53 \\
& SA-GATE~\cite{Chen2020} & 97.41 & 86.80 & 86.68 & 95.68 & 92.60 & 90.21 & 91.40 & 84.53 \\
& CMGFNet~\cite{Hosseinpour2022} & 96.63 & 82.65 & 89.98 & 93.17 & 91.93 & 90.01 & 90.97 & 83.74 \\
& TransUNet~\cite{Chen2024} & 97.63 & 87.40 & 88.00 & 95.68 & 92.84 & 91.16 & 92.10 & 85.63 \\
& CMFNet~\cite{Ma2022} & 97.23 & 84.29 & 89.03 & 95.49 & 91.62 & 90.22 & 91.26 & 84.26 \\
& MFTransNet~\cite{He2023a} & 97.37 & 85.71 & 86.92 & 96.05 & 92.45 & 89.96 & 90.00 & 84.04 \\
& FTransUNet~\cite{Ma2024} & \textbf{97.78} & 88.27 & 88.48 & 96.31 & 93.17 & 91.34 & 92.41 & 86.2 \\
\multirow{-10}{*}{Multiple} & \cellcolor[HTML]{D3D3D3}{AMMNet (Ours)} & \cellcolor[HTML]{D3D3D3}{94.78} & \cellcolor[HTML]{D3D3D3}{\textbf{97.47}} & \cellcolor[HTML]{D3D3D3}{\textbf{90.93}} & \cellcolor[HTML]{D3D3D3}{87.47} & \cellcolor[HTML]{D3D3D3}{\textbf{96.69}} & \cellcolor[HTML]{D3D3D3}{\textbf{93.47}} & \cellcolor[HTML]{D3D3D3}{\textbf{93.50}} & \cellcolor[HTML]{D3D3D3}{\textbf{88.02}} \\ \hline
\end{tabular}%
}
\end{table*}

Compared to Vaihingen, which is a small village characterized by detached houses and low-rise buildings, Potsdam represents a more complex urban environment with large building blocks and densely built-up areas.

In Potsdam, the ``Tree'' category is relatively sparse and more challenging to identify due to the limited number of training samples.
As shown in Table~\ref{tab:result_potsdam}, both single-modality and multi-modality networks exhibit limited performance in this class compared to Vaihingen.
Nevertheless, AMMNet demonstrates superior recognition capability, achieving a 9.20\% improvement compared with the previous SOTA network, which highlights its effectiveness in detecting sparse categories by leveraging structural cues from DSM.
The performance drop observed in the ``Building'' and ``Car'' classes may be attributed to the reduced utility of DSM information in dense urban areas, where elevation cues are ambiguous due to structural uniformity and occlusions.

Despite these class-level variations, AMMNet maintains strong overall performance on the Potsdam dataset, achieving an mOA of 93.47\%, mF1 of 93.50\%, and mIoU of 88.02\%, consistently outperforming both single-modality and other multi-modality networks.

\subsection{Qualitative Analysis}
This section presents a qualitative analysis of segmentation results produced by two single-modal networks (Swin-UNet and UNetformer) and three multi-modal networks (MFTransNet, FTransUNet, and the proposed AMMNet).

\begin{figure*}
    \centering
    \includegraphics[width=\textwidth]{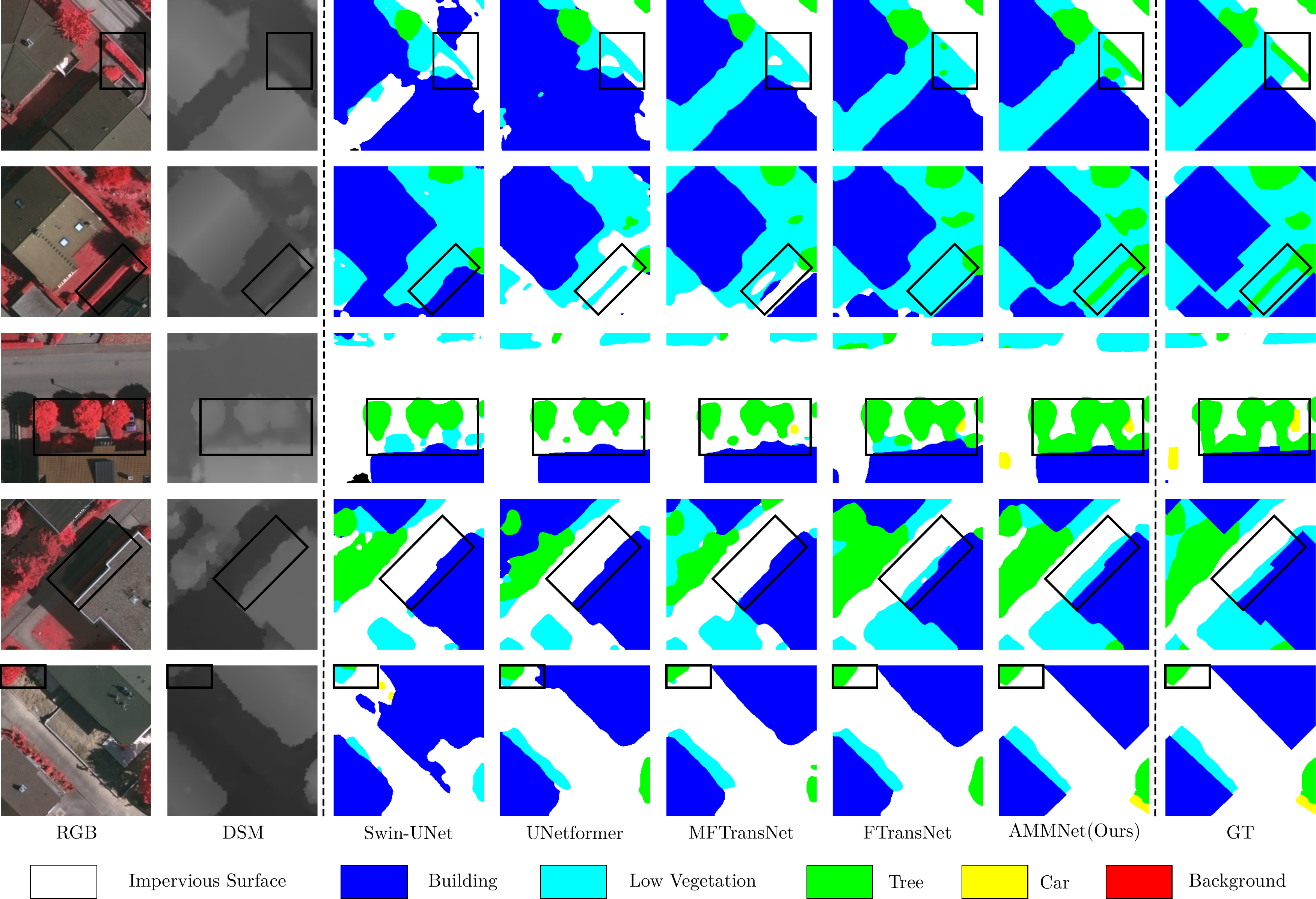}
    \caption{Qualitative comparison of segmentation results on the ISPRS Vaihingen test set. The figure highlights AMMNet’s superior ability to accurately segment occluded and contextually ambiguous classes, such as “Tree” and “Low Vegetation”.}
    \label{fig:vaihingen_vis}
\end{figure*}

Fig.~\ref{fig:vaihingen_vis} illustrates the comparative performance on the ISPRS Vaihingen test set.
The results show that single-modal networks struggle with objects affected by occlusion, likely because occlusions introduce ambiguity into the RGB-based contextual features.
On the other hand, existing multi-modal networks are able to identify occluded objects, however, they often confuse the ``Tree'' and ``Low Vegetation'' classes, which is attributed to their suboptimal fusion strategies.

In comparison, AMMNet demonstrates more accurate segmentation on occluded and contextually ambiguous classes with its effective integration of RGB and DSM features via its asymmetric designs.

\begin{figure*}
    \centering
    \includegraphics[width=\textwidth]{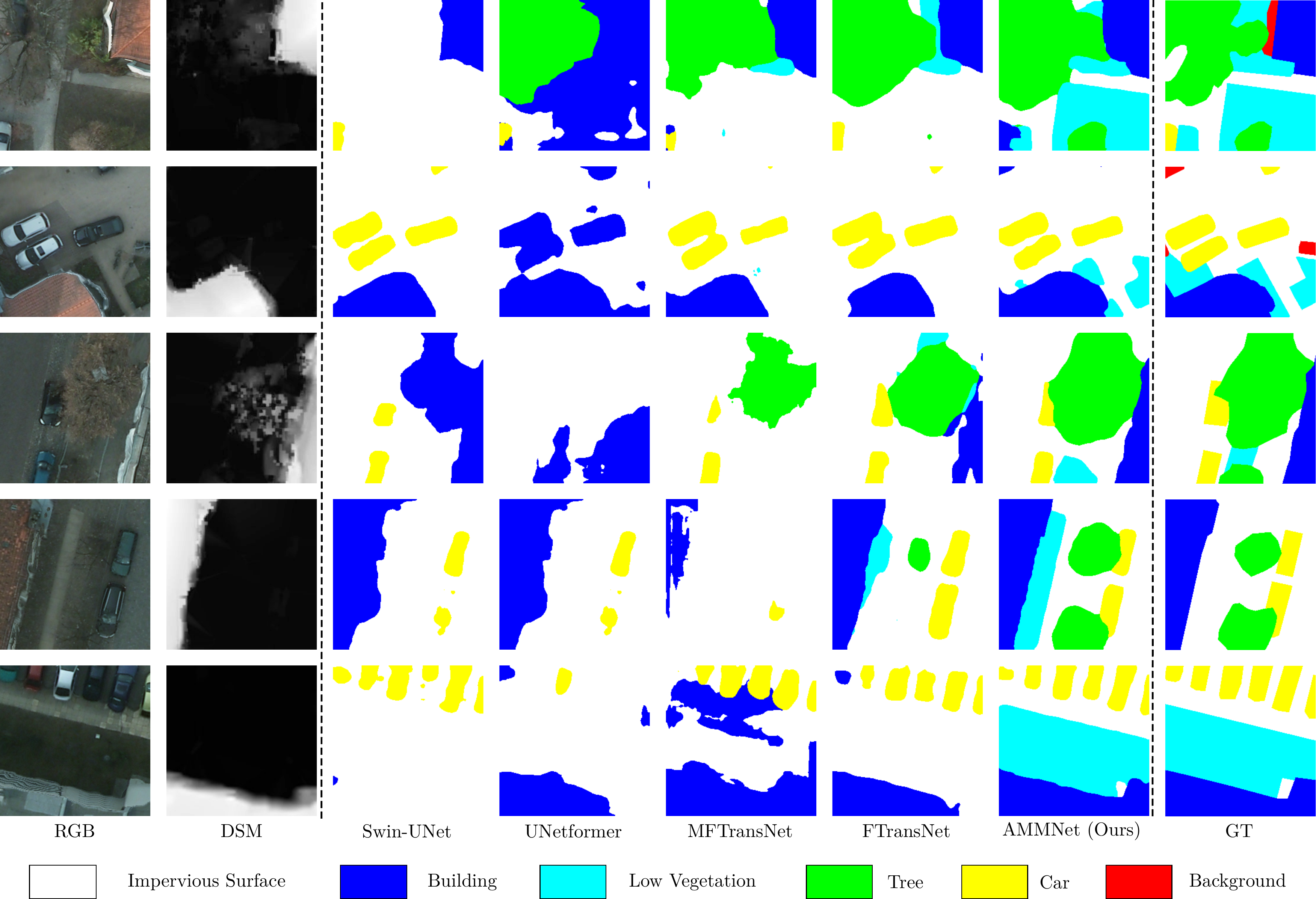}
    \caption{Qualitative comparison of segmentation result on the ISPRS Potsdam test set (input size: $256 \times 256$). Single-modal networks (Swin-UNet and UNetformer) exhibit limited qualitative performance. Multi-modal networks (MFTransNet and FTransNet) show improved performance but struggle with fine-grained classes such as ``Low Vegetation''. The proposed AMMNet achieves the most accurate results, successfully identifying the majority of objects.}
    \label{fig:potsdam_vis}
\end{figure*}

Compared with Vaihingen, the Potsdam dataset presents a more challenging scenario due to its dense and complex urban environment.
Fig.~\ref{fig:potsdam_vis} presents a qualitative comparison of segmentation results on the ISPRS Potsdam test set.
As illustrated, Swin-UNet has difficulty distinguishing certain objects due to contextual similarity, while UNetformer not only fails to detect some instances but also produces incorrect semantic predictions, particularly in the ``Car'' category. 

Among the multi-modal networks, MFTransNet and FTransNet produce more accurate segmentation results overall compared to single-modal methods.
However, they fail to correctly identify certain ``Low Vegetation'' regions, likely due to suboptimal integration of structural information from the DSM modality.

Among all the compared networks, AMMNet delivers the most accurate qualitative results, successfully identifying the majority of objects with high consistency. 
Its better qualitative result highlights the effectiveness of the proposed asymmetric architecture in integrating RGB and DSM information.

\subsection{Efficiency Analysis}
Table~\ref{tab:complexity_analysis} presents a comprehensive comparison of state-of-the-art segmentation networks in terms of model efficiency. The efficiency metrics include FLOPs, parameter count, and memory usage, while mIoU is provided as the segmentation performance metric.

As shown in the table, single-modal networks exhibit lower computation complexity but yield inferior segmentation accuracy compared to multi-modal counterparts.
This trade-off is expected, as reduced computational complexity often correlates with limited model capacity, which restricts the network's ability to learn rich representations.
Furthermore, the use of only a single modality limits the diversity of information accessible to the model, compounding the performance degradation.

In general, multi-modal networks achieve higher segmentation accuracy by leveraging additional modalities and increased model capacity.
The incorporation of complementary information, such as contextual and structural information, contributes to this performance gain.
Notably, even with increased FLOPs, parameter count and memory usage, some multi-modal networks (vFuseNet, FuseNet, ESANet, and TransUNet) still underperform the best-performing single-modal counterpart (UNetFormer), which highlights the importance of effective exploitation of cross-modal information.

Standing out among multi-modal networks, the proposed AMMNet demonstrates an exceptional performance–efficiency trade-off, achieving the highest segmentation accuracy while requiring only 64\% of the FLOPs, 30\% of the memory consumption, and a comparable number of parameters compared to the second-best method, FTransfUNet.
%

The substantial reduction in computational complexity is primarily attributed to the Asymmetric Dual Encoder (ADE), while the performance improvement results from alleviating misalignment through the Asymmetric Prior Fusion (APF) and Dual Alignment (DA) modules.
These results demonstrate that the asymmetric design enables AMMNet to serve as an efficient and performant multi-modal segmentation network.

\begin{table}[!t]
\centering
\caption{Comparison of model efficiency. The best results are highlighted in bold.}
\label{tab:complexity_analysis}
\resizebox{\linewidth}{!}{%
\begin{tabular}{c|c|ccc|c}
\hline
\makecell{\textbf{Modality}} & \makecell{\textbf{Models}} & \makecell{\textbf{FLOPs} \\ (G) $\downarrow$} & \makecell{\textbf{Param.} \\ (M) $\downarrow$} & \makecell{\textbf{Mem.} \\ (MB) $\downarrow$} & \makecell{\textbf{mIoU} \\ (\%) $\uparrow$} \\
\hline
& ABCNet~\cite{Liu2021} & \textbf{3.90} & 13.39 & 1598 & 75.20 \\
& PSPNet~\cite{Zhao2017} & 49.03 & 46.72 & 3124 & 76.96 \\
& MAResU-Net~\cite{Li2022} & 8.79 & 26.27 & 1908 & 79.89 \\
& Swin-UNet~\cite{Cao2023} & 16.54 & 34.68 & 1297 & 76.03 \\
\multirow{-5}{*}{Single} & UNetFormer~\cite{Wang2022} & 6.04 & \textbf{24.20} & 1980 & 81.97 \\ \hline
& vFuseNet~\cite{Audebert2018} & 60.36 & 44.17 & 2618 & 78.92 \\
& FuseNet~\cite{Hazirbas2017} & 58.37 & 42.08 & 2284 & 78.71 \\
& ESANet~\cite{Seichter2022} & 7.73 & 34.03 & 1914 & 79.42 \\
& TransUNet~\cite{Chen2024} & 32.27 & 93.23 & 3028 & 78.26 \\
& SA-GATE~\cite{Chen2020} & 41.28 & 110.85 & 3174 & 81.27 \\
& CMFNet~\cite{Ma2022} & 78.25 & 123.63 & 4058 & 81.44 \\
& MFTransNet~\cite{He2023a} & 8.44 & 43.77 & 1549 & 81.61 \\
& CMGFNet~\cite{Hosseinpour2022} & 19.51 & 64.20 & 2463 & 82.26 \\
& FTransUNet~\cite{Ma2024} & 45.21 & 160.88 & 3463 & 84.23 \\
\multirow{-10}{*}{Multiple} & \cellcolor[HTML]{D3D3D3}{AMMNet (Ours)} & \cellcolor[HTML]{D3D3D3}{28.82} & \cellcolor[HTML]{D3D3D3}{151.26} & \cellcolor[HTML]{D3D3D3}{\textbf{1026}} & \cellcolor[HTML]{D3D3D3}{\textbf{87.56}} \\ \hline
\end{tabular}%
}
\end{table}

\subsection{Ablation Study}
This section first evaluates the individual contributions and interdependencies of the core components in AMMNet.
Next, we investigate the effectiveness of the asymmetric design in ADE for the RGB-DSM modality pair through a series of controlled experiments.
Finally, we assess the impact of the DA module's alignment strength on segmentation performance.

\subsubsection{Component Contribution}

\begin{table}[!t]
\tiny
\centering
\caption{Ablation study of each component in AMMNet. Best values are highlighted in Bold.}
\label{tab:ablation_component}
\resizebox{.9\linewidth}{!}{%
\begin{tabular}{ccc|ccc}
\hline
\textbf{APF} & \textbf{DA} & \textbf{ADE} & \textbf{mOA} $\uparrow$ & \textbf{mF1} $\uparrow$ & \textbf{mIoU} $\uparrow$ \\ \hline
\xmark & \xmark & \xmark & 92.72 & 92.57 & 86.33 \\ \hline
\cmark & \xmark & \xmark & 92.65 & 92.63 & 86.44 \\
\xmark & \cmark & \xmark & 92.77 & 92.52 & 86.29 \\
\xmark & \xmark & \cmark & 92.92 & 92.58 & 86.35 \\ \hline 
\xmark & \cmark & \cmark & 92.69 & 92.59 & 86.37 \\
\cmark & \xmark & \cmark & 92.99 & 93.02 & 87.12 \\
\cmark & \cmark & \xmark & 93.33 & 93.12 & 87.33 \\ \hline
\cellcolor[HTML]{D3D3D3}{\cmark} & \cellcolor[HTML]{D3D3D3}{\cmark} & \cellcolor[HTML]{D3D3D3}{\cmark} & \cellcolor[HTML]{D3D3D3}{\textbf{93.52}} & \cellcolor[HTML]{D3D3D3}{\textbf{93.27}} & \cellcolor[HTML]{D3D3D3}{\textbf{87.56}} \\ \hline
\end{tabular}%
}
\end{table}

To assess the individual contribution and interdependency of each module, a series of ablation baselines is constructed. 
When ADE is disabled, a symmetrical dual encoder architecture is adopted, where both branches use the same encoder variant as the RGB branch in ADE.
To evaluate the impact of the APF module, it is replaced with a naive fusion strategy consisting of feature concatenation followed by a $1 \times 1$ convolution.
For the DA module, its effect is isolated by setting the alignment loss weight $\alpha$ in Eq.~\ref{eq:final_loss} to zero, thereby disabling the alignment constraint.

As shown in Table~\ref{tab:ablation_component}, applying APF and ADE individually yields marginal performance gains compared to the baseline with all components disabled, demonstrating their effectiveness as standalone modules.
Moreover, the combination of APF and ADE leads to additional gains, highlighting the effectiveness of the proposed asymmetric design in enhancing representational learning.

In contrast, the effectiveness of DA appears to depend on the presence of APF. Specifically, adding DA alone, or in combination with ADE, yields limited or even degraded performance. However, when DA is used alongside APF, substantial improvements are observed across all three metrics: mOA increases by 0.61\% to 93.33\%, mF1 by 0.55\% to 93.12\%, and mIoU by 1.00\% to 87.33\%. These results indicate that DA’s distributional alignment is most beneficial when APF is available to facilitate effective modality-aware fusion and encourage feature complementarity. In the absence of APF, DA introduces adverse effects, likely due to its undesired intervention with irrelevant information.

Finally, the combined application of all three components yields the best performance, with a notable 0.80\% increase in mOA, 0.70\% in mF1, and 1.23\% in mIoU over the baseline with all proposed modules disabled.

\subsubsection{Asymmetric Encoder Design}
To evaluate the effectiveness of the asymmetric design in ADE, a series of experiments are conducted using different encoder configurations.
The results are provided in Table~\ref{tab:encoder_variant}.

\begin{table}[!t]
\tiny
\centering
\caption{Performance comparison of ADE configurations. Best values are highlighted in bold.}
\label{tab:encoder_variant}
\resizebox{.9\linewidth}{!}{%
\begin{tabular}{cc|ccc}
\hline

\textbf{RGB} & \textbf{DSM} & \textbf{mOA} $\uparrow$ & \textbf{mF1} $\uparrow$ & \textbf{mIoU} $\uparrow$ \\ \hline
Tiny & Tiny & 91.97 & 91.76 & 85.04 \\
Tiny & Small & 91.73 & 91.90 & 85.25 \\
Tiny & Base & 91.34 & 91.79 & 85.09 \\ \hline
Small & Tiny & 92.03 & 91.78 & 85.03 \\
Small & Small & 92.03 & 91.66 & 84.87 \\
Small & Base & 91.77 & 91.76 & 85.00 \\ \hline
Base & Tiny & 92.77 & 92.88 & 86.94 \\ 
\cellcolor[HTML]{D3D3D3}{Base} & \cellcolor[HTML]{D3D3D3}{Small} & \cellcolor[HTML]{D3D3D3}{\textbf{93.52}} & \cellcolor[HTML]{D3D3D3}{\textbf{93.27}} & \cellcolor[HTML]{D3D3D3}{\textbf{87.56}} \\
Base & Base & 92.87 & 92.95 & 87.03 \\ \hline
\end{tabular}%
}
\end{table}

The results demonstrate that larger encoder variants for both modalities generally lead to improved segmentation performance, highlighting the importance of representational capacity in encoder design.
Nevertheless, the optimal configuration is achieved by using the ``Base'' variant for RGB and the ``Small'' variant for DSM.
This finding validates the effectiveness of the asymmetric design, which strategically assigns greater representational capacity to the RGB modality to fully exploit its rich contextual content while assigning reduced capacity to the DSM modality, which primarily encodes sparse structural cues.
Furthermore, configurations in which the RGB encoder is deeper than the DSM encoder consistently outperform the inverse setups, reinforcing the importance of modality-specific encoder design tailored to each modality.

\subsubsection{Strength Sensitivity of Distribution Alignment}
Table~\ref{tab:kld weight} demonstrates the strength sensitivity of DA by varying the alignment loss weight $\alpha$. 

\begin{table}[!t]
\tiny
\centering
\caption{Ablation study of alignment loss weight ($\alpha$) in AMMNet. Best values are highlighted in Bold.}
\label{tab:kld weight}
\resizebox{0.75\linewidth}{!}{%
\begin{tabular}{c|ccc}
\hline
$\alpha$ & \textbf{mOA} $\uparrow$ & \textbf{mF1} $\uparrow$ & \textbf{mIoU} $\uparrow$ \\ \hline
$1\mathrm{e}{-3}$ & 93.11 & 92.98 & 87.06 \\
$75\mathrm{e}{-5}$ & 93.07 & 92.86 & 86.95 \\
\cellcolor[HTML]{D3D3D3}{$5\mathrm{e}{-4}$} & \cellcolor[HTML]{D3D3D3}{\textbf{93.52}} & \cellcolor[HTML]{D3D3D3}{\textbf{93.27}} & \cellcolor[HTML]{D3D3D3}{\textbf{87.56}} \\
$25\mathrm{e}{-5}$ & 93.09 & 92.89 & 87.00 \\
$1\mathrm{e}{-4}$ & 91.46 & 92.25 & 86.41 \\ \hline
\end{tabular}%
}
\end{table}

It can be observed that when $\alpha$ is set too low, AMMNet exhibits suboptimal performance, as the DA fails to suppress irrelevant information effectively.
Conversely, an excessively large $\alpha$ shifts optimization focus away from semantic accuracy toward minimizing divergence between modalities, potentially degrading task performance.

The best performance is obtained with $\alpha = 5\mathrm{e}{-4}$, which achieves a favorable trade-off between distributional alignment and supervised learning.
This balance maximizes the utility of the DA module, enabling robust and efficient multi-modal integration.

\section{Conclusion}
This paper identifies the complementary characteristics of RGB and DSM inputs and highlights two key limitations in multi-modality semantic segmentation for remote sensing: architectural redundancy and modality misalignment.
To address these issues, we propose a novel segmentation model with asymmetric architecture, termed Asymmetric Multi-Modal Network (AMMNet), which tackles these challenges through three dedicated modules.
To mitigate architectural redundancy, AMMNet incorporates an Asymmetric Dual Encoder (ADE), which enables modality-specific feature extraction with reduced computational overhead.
To resolve modality misalignment, an Asymmetric Prior Fuser (APF) constructs structure-aware contextual features by fusing RGB and DSM representations through a modality-aware prior matrix. 
Additionally, a Distribution Alignment (DA) module performs distribution-level alignment by minimizing divergence between the feature distributions, preserving complementary information while suppressing irrelevant information.
This asymmetric design enables AMMNet to achieve robust segmentation performance in complex urban environments, as validated by the superior results on the ISPRS Vaihingen and Potsdam datasets across three key evaluation metrics.

\bibliographystyle{IEEEtran}
\bibliography{AMMNet.bib}

\end{document}